 \documentclass{vgtc}                          

\ifpdf
\pdfoutput=1\relax                   
\usepackage{graphicx}                
\DeclareGraphicsExtensions{.pdf,.png,.jpg,.jpeg} 
\else
\usepackage{graphicx}                
\DeclareGraphicsExtensions{.eps}     
\fi%

\graphicspath{{figures/}{pictures/}{images/}{./}} 

\DeclareMathAlphabet{\mathcal}{OMS}{cmsy}{m}{n} 
\usepackage{amsmath}

\usepackage{textcomp}
\usepackage{stfloats}
\usepackage{url}
\usepackage{verbatim}
\usepackage{graphicx}
\usepackage{cite}
\usepackage{bm}
\usepackage{color}
\usepackage{multirow}

\usepackage{microtype}                 
\PassOptionsToPackage{warn}{textcomp}  
\usepackage{textcomp}                  
\usepackage{mathptmx}                  
\usepackage{times}                     
\usepackage{cite}                      
\usepackage{tabu}                      
\usepackage{booktabs}                  

\usepackage{stfloats}
\usepackage{url}
\usepackage{verbatim}
\usepackage{graphicx}
\usepackage{cite}
\usepackage{bm}
\usepackage{color}
\usepackage{multirow}

\newcommand{\equref}[1]{Equ.~(\ref{#1})}
\newcommand{\figref}[1]{Fig.~\ref{#1}}

\newcommand{\tabref}[1]{Tab.~\ref{#1}}

\DeclareMathOperator*{\argmin}{arg\,min}
\renewcommand{\vec}[1]{\boldsymbol{#1}}

\newcommand{\T}{\mathcal{T}}
\newcommand{\F}{\mathcal{F}}
\newcommand{\x}{\vec{x}}

\onlineid{0}

\vgtccategory{Research}

\vgtcinsertpkg

\title{CompenHR: Efficient Full Compensation for High-resolution Projector}

\author{Yuxi Wang\thanks{e-mail: yxwang@hdu.edu.cn}\\ %
	\scriptsize Hangzhou Dianzi University %
	\and Haibin Ling\thanks{e-mail: hling@cs.stonybrook.edu}\\ %
	\scriptsize Stony Brook University %
	\and Bingyao Huang\thanks{e-mail: bhuang@swu.edu.cn. Corresponding author}\\ %
	\scriptsize Southwest University }

\abstract{
Full projector compensation is a practical task of projector-camera systems. It aims to find a projector input image, named compensation image, such that when projected it cancels the geometric and photometric distortions due to the physical environment and hardware. State-of-the-art methods use deep learning to address this problem and show promising performance for low-resolution setups. However, directly applying deep learning to high-resolution setups is impractical due to the long training time and high memory cost. To address this issue, this paper proposes a practical full compensation solution. Firstly, we design an attention-based grid refinement network to improve geometric correction quality. Secondly, we integrate a novel sampling scheme into an end-to-end compensation network to alleviate computation and introduce attention blocks to preserve key features. Finally, we construct a benchmark dataset for high-resolution projector full compensation. In experiments, our method demonstrates clear advantages in both efficiency and quality. Our code is available at \url{https://github.com/cyxwang/CompenHR}.
} 

\CCScatlist{
\CCScatTwelve{Projector compensation}; {Spatial augmented reality}; {Projector-camera system}
}

\begin{document}

	
	
	\maketitle
	
	\section{Introduction} 
	
	As an essential device for spatial augmented reality, projectors are usually combined with cameras to form smart projector-camera systems, and {are} used in many scientific experiments and real-world applications\cite{Raskar_2000, Bimber_2006,Yoshida_2010,Kemmoku_2006, Ehnes_2006,Bimber_2005,Low_2003,Kenyon_2014, Ozacar_2015,Majumder_2015,Krum_2016,Narita_TVCG_2017,Hiratani_2019,Ueda_2020,Punpongsanon_TVCG_2020}. 
	However, projection onto non-planar and textured surfaces is still a challenging problem, which limits the applicability of projector-camera systems.
	As a typical solution, full projector compensation neutralizes {geometric and photometric} distortions caused by sensor radiometric variation, lens distortion, and surface material reflectance~\cite{Asayama_TVCG_2018,Harville_CVPRW_2006,Siegl_ACMTG_2015,Raskar_CVPR_2001,Tehrani_TVCG_2021,Yu_TIP_2021,Shih_TIP_2021,Grundhofer_TIP_2015,Fujii_CVPR_2005,Huang_TVCG_2021,Miyagawa_TIP_2014,Huang_TASE_2021}. 
	In particular, a composite function of full projector compensation is estimated from projector input and the corresponding camera-captured images, and then, the compensation image is generated based on the estimated parameters.
	
	
	Traditional projector compensation methods assume that geometric and photometric distortions are independent. Thus, they formulate these two tasks separately. For geometric correction, a common solution is finding the pixel-to-pixel correspondences with structured light, then generating the corrected image by inverse mapping. For photometric compensation, conventional methods define a per-pixel color mapping function for each camera and projector pixel pair.
	Recently, with the successful application of deep learning, some works are devoted to modeling full compensation using deep neural networks\cite{Huang_ICCV_2019,Huang_PAMI_2021}. Although these end-to-end algorithms overcome the drawbacks of two-step methods, the {memory usage and} computation cost increase rapidly with image resolutions, and thus they are less practical for high-resolution setups. Therefore, how to compensate for the high-resolution input of a projector under constrained conditions has yet to be studied.
	\begin{figure}[t]
		\includegraphics[width=1.0\columnwidth]{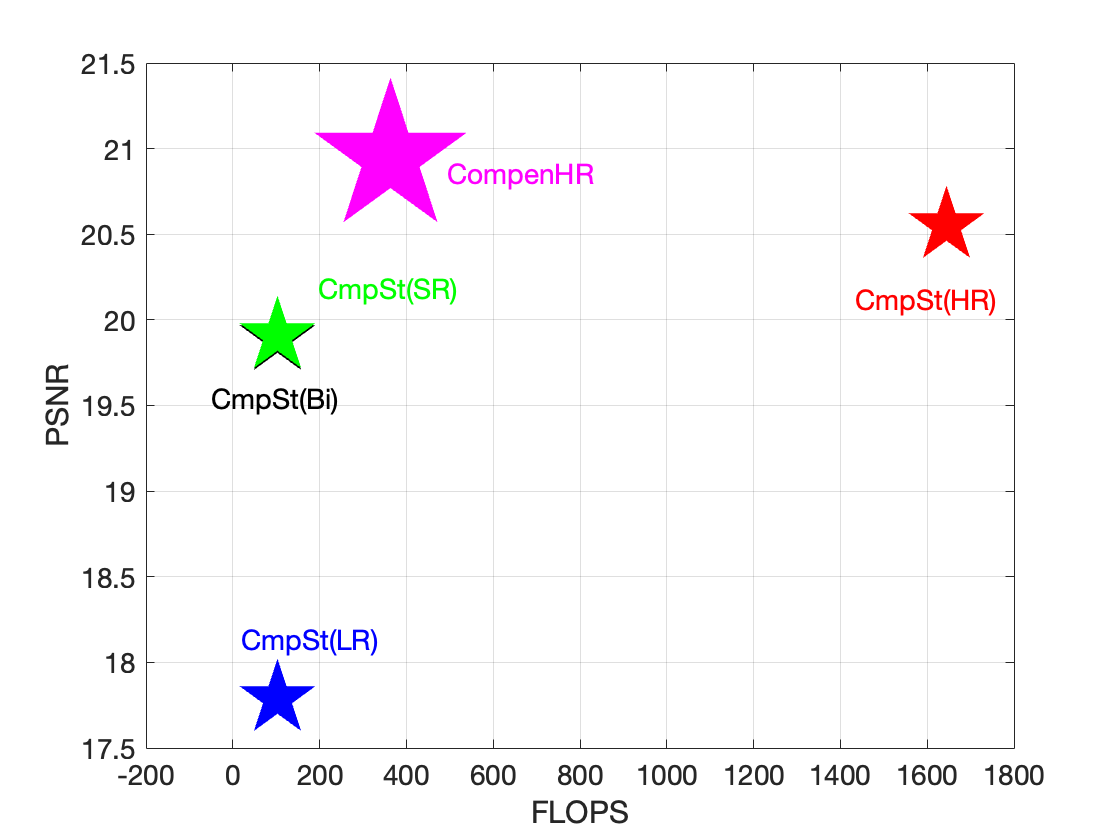} 
		\caption{Comparison of state-of-the-art end-to-end full compensation algorithms. FLOPS are calculated on an Nvidia GeForce 1080 GPU with input size $1024\times 1024\times 3$. The star sizes are proportional to the number of parameters. The proposed CompenHR (the magenta star) achieves the highest PSNR with a mediate FLOPS. Note that CmpSt(Bi) and CmpSt(SR) are highly overlapped.}		
		\centering
		\vspace{-1.0em}
		\label{example}		
	\end{figure}
	
	
	By taking into account the memory and time limitations, this paper builds an efficient end-to-end trainable framework named CompenHR for full high-resolution projector compensation.  
	After reformulating the problem by integrating a sampling scheme, we design efficient subnets to model compensation functions. For geometric correction, a novel subnet named GANet (short for Attention-based Geometry Correction Network) is utilized to warp high-resolution input with corrected geometry. We design a grid refinement network to improve the accuracy of sampling grid estimation. Then, for photometric compensation, an efficient subnet named PANet (short for Attention-based Photometric Compensation Network) is exploited to recover the high-resolution images. We employ shuffle/unshuffle\cite{Shi_CVPR_2016}, instead of traditional downsample/upsample, to improve PANet and enable it to be trained more efficiently with a small amount of information loss. Moreover, we integrate the attention mechanism into both subnets and thus allow CompenHR to extract more important features from the high-resolution input.
	In addition, due to the lack of high-resolution compensation datasets for evaluation, we construct a dataset with 24 different projector-camera system setups. In experiments, the proposed CompenHR is clearly more efficient than state-of-the-art methods.	
	Our contributions can be summarized as follows: %
	\begin{itemize}
		\vspace{-2mm}\item
		{We reformulate the full compensation problem for high-resolution projectors and propose a memory and time efficient solution named CompenHR.}
		\vspace{-2mm}\item {We design an efficient sampling grid refinement subnet for geometric correction, owing to which our CompenHR can achieve more accurate image warping than the state-of-the-art methods.}
		\vspace{-2mm}\item  {Instead of simple downsampling/upsampling, we apply novel pixel unshuffle/shuffle operations in photometric compensation. Such a design not only avoids information loss but also improves network training and inference efficiency. Moreover, the pixel attention mechanism is integrated into both subnets to focus on key features, which further improves our CompenHR performance.}
		\vspace{-2mm}\item {A {real} high-resolution projector full compensation benchmark dataset {with 25 setups} is constructed and is expected to facilitate future work in this direction. {In addition, a synthetic high-resolution dataset with 100 setups is proposed to pre-train models.}}
	\end{itemize}
	\section{Related works} 
	\subsection{Compensation methods}
	Projector compensation is an important task for spatial augmented reality, and it has been studied extensively. Existing methods can be divided into three categories: geometry correction, photometric compensation, and full compensation. Detailed reviews can be found in \cite{Bimber_2008, grundhofer2018recent}.
	\subsubsection{Geometric correction}
	For conventional applications, where projection surfaces are planar or multi-planar, traditional methods estimate geometric relations between the camera, the projector, and the projection surface. Projections can be simply corrected by homographies~\cite{Raskar_2000,Raskar_2001}. 
	However, curved surfaces increase the intricacy of geometric correction in many applications. 
	
	A surge of work estimates the pixel mappings between the projector input and camera-captured images using structured light~\cite{Raskar_2003,Tardif_3DIM_2003,Boroomand_ICIP_2016,Lin_PR_2016,Willi_ISMAR_2017}. These methods project landmarks onto the surface and capture them with a synchronized camera. Then the 3D geometry of the surface is reconstructed given the pixel mappings and the geometric relationships between the cameras, the projectors, and the surfaces. 
	To reduce the computational complexity, Boroomand \emph{et al}.~\cite{Boroomand_ICIP_2016} propose a geometric correction method based on local surface saliency that selects a small set of points rather than dense samples.
	Tardif \emph{et al}.\cite{Tardif_3DIM_2003} decompose the mapping function from the camera to the projector into two orientations and determine its parameters by the correspondence of each pixel without surface reconstruction, then construct the corrected image by inverse mapping.
	Tehrani \emph{et al}.\cite{Tehrani_TVCG_2021} study an automatic method to estimate all device parameters and the surface geometry for a multi-projector system without prior calibration. 
	Particularly, some efforts track the dynamic non-planar surface by marking patterns with invariant topologies~\cite{Guskov_ICPR_2002,Narita_TVCG_2017}.
	Narita \emph{et al}.~\cite{Narita_TVCG_2017} design fiducial markers that consist of four types of dot clusters, and track non-rigid surfaces by identifying these dot cluster IDs in real-time.
	\subsubsection{Photometric compensation}
	Photometric compensation aims to cancel the photometric distortion caused by the textured projection surface and the radiometric response functions, with the assumption that captured images have already been geometrically corrected.
	Previous methods estimate the color transformation by 1-to-1 mapping from the camera to the projector pixels. 
	Nayar \emph{et al}.\cite{Nayar_2003} define the mapping function with a $3\times3$ color mixing matrix and estimate it using the correspondence between the captured image and the projected image. On this basis, Grossberg \emph{et al}. \cite{Grossberg_CVPR_2004} reduce the number of calibration patterns to six. 
	Grundhöfer and Iwai.\cite{ Grundhofer_CVPRW_2013,Grundhofer_TIP_2015} propose a method for an uncalibrated projector and camera system. The compensation process is modeled by a non-linear color mapping function that is defined by a per-pixel thin plate spline interpolation. Considering the pixel redundancies of surface reflectance and the input coherence of the transfer function, Li \emph{et al}.\cite{Li_CGF_2018} employ sparse sampling and multidimensional interpolation techniques to improve compensation efficiency. 
	
	However, the limitation of dynamic ranges and gamuts of the projector and camera system results in clipping artifacts in compensation images. To address this issue, some studies take human vision system properties into consideration. For instance, Wang \emph{et al}. \cite{Wang_CVPR_2005} employ the perceptually-based physical error metric, which incorporates the threshold sensitivity, contrast sensitivity, and visual mask to minimize achromatic artifacts in compensation images. 
	Huang \emph{et al}.\cite{Huang_TIP_2017} adjust the brightness and hue of the image by manipulating the reference white of the CIECAM02 Color Appearance Model according to the anchoring property. Pjanic \emph{et al}.\cite{Pjanic_TVCG_2018}  propose an adaptive color gamut acquisition to generate a color-prediction model, and then optimize the framework in the RLab color space. Akiyama \emph{et al}.\cite{Akiyama_VRW_2022} generate the compensation image by minimizing the perceptual distance between its projection and the desired image.

	Besides, for a dynamic environment, Fujii \emph{et al}.\cite{Fujii_CVPR_2005} present an adaptive photometric model under the assumption that 
	the global light is approximately unchanged. 
	In their method, parameters are first estimated by projecting four uniform calibration images, and then the surface reflectance matrix is updated using the error between the captured and desired images when the surface reflectance change exceeds the threshold.
	Bokaris \emph{et al}.\cite{Bokaris_ECCV_2014,Bokaris_ICIP_2015} generate images using a linear transformation matrix for dynamic surfaces with one-frame delay. Hashimoto \emph{et al}.\cite{Hashimoto_TVC_2021} estimate the offset of the adjacent moment to update the inter-pixel correspondence and optimize the current reflectance using the present and the sum of past correspondence. 
	Considering the effect of inter-pixel coupling, Shih \emph{et al}.\cite{Shih_TIP_2021} calculate the gamma function and the inter-pixel coupling matrix using two constant grayscale patterns and a ramp grayscale one in the initial calibration, then estimate the dynamic reflectance using the projected image as calibration patterns.  
	
	Inspired by the successful application of deep learning to image-to-image translation tasks, Huang \emph{et al}. \cite{Huang_CVPR_2019} explore an end-to-end photometric compensation method that learns the inverse mapping from the camera image to the projector image using convolution neural networks and achieves outstanding performance {in static projector-camera systems. For white diffuse surfaces, Kageyama \emph{et al}. \cite{Kageyama_TVCG_2022} propose an effective deblurring technique using a convolutional neural network for dynamic projection mapping scenarios. It employs an extractor to estimate defocus blur and luminance attenuation maps and then feeds them to a generator to compute compensation images.} 
	
	\subsubsection{Full compensation}
	Full compensation techniques perform geometry correction and photometric compensation jointly. 
	Park \emph{et al}.\cite{Park_TCSVT_2008} present spatial and temporal encoding techniques that compensate images via embedding patterns. In temporal encoding, pattern images for geometric and radiometric calibration are projected and embedded separately; in spatial encoding, a single pattern that incorporates the information for both geometric and radiometric calibration is designed for simultaneous compensation.
	Shahpaski \emph{et al}.\cite{Shahpaski_CVPR_2017} also design a special projected pattern using squares with mixed blue and red colors for geometric and radiometric calibration. They project this special pattern onto a printed pattern with a standard checkerboard. Benefiting from this design,  printed and projected corners are able to be detected from the blue channel and the red channel of captured images respectively using automatic checkerboard detectors.
	
	Recently, Huang \emph{et al}.\cite{Huang_ICCV_2019,Huang_PAMI_2021} reformulate the physical process of full compensation and learn the geometric correction and photometric compensation functions using deep neural networks. Park \emph{et al}.\cite{Park_2022} simulate the full projection process under virtual light and optimize the compensation image using differentiable rendering.
	\subsection{Our method}
	Our method, named CompenHR, belongs to the category of full compensation and is inspired by CompenNeSt++~\cite{Huang_PAMI_2021}. While CompenNeSt++ has achieved promising performance on low-resolution setups, its memory and training time grow dramatically with the increase of image sizes, making it impractical to compensate for high-resolution inputs. To address this issue, we reformulate the full compensation process for high-resolution projectors and propose to reduce the feature map sizes in the photometric compensation module. After that, we design networks by combining a variety of effective schemes to further improve accuracy. 
	For geometric correction, an attention-based network is designed to refine the sampling grid, produces accurate image warping; for photometric compensation, novel sampling operations are introduced to rearrange images and feature maps. Furthermore, attention mechanisms are employed to preserve key features from images and their linear transformations. 
	Benefiting from these schemes, our CompenHR shows great advantages in memory and time efficiency, with even slightly improved projection quality. 
	
	\subsection{{Attention Mechanism}}
	The attention mechanism used in our CompenHR is inspired by its popularity in computer vision. In the following, we discuss some most related works.
	A pioneer channel-wise attention mechanism is the squeeze-and-excitation (SE) module \cite{Hu_CVPR_2018}, which emphasizes the channels with key information of feature maps. On this basis, Hui \emph{et al}. \cite{Hui_ACMMM_2019} construct the contrast-aware channel attention block by replacing the pooling with a contrast operation.
	The Squeeze-and-Attention (SA) module\cite{Zhong_CVPR_2020} replaces the full convolutional layers in SE with the pooling and upsampling operations. 
	Zhang \emph{et al}. \cite{Zhang_ECCV_2018} design the residual channel attention network (RCAN) using residual channel attention blocks. 
	Dai \emph{et al}. \cite{Dai_CVPR_2019} propose the second-order attention network (SAN) by considering the high-order channel feature correlations.
	
	Additionally, a surge of methods incorporate both channel-wise attention and spatial attention.
	Features in Long \emph{et al.}\cite{Long_CVPR_2017} are weighted by the cascaded channel-wise attention and spatial attention modules. 
	Woo \emph{et al}.\cite{Woo_ECCV_2018} arrange the channel-wise and spatial attention modules in parallel and sequentially respectively.
	Liu \emph{et al}.\cite{Liu_CVPR_2020} propose the enhanced spatial attention (ESA) blocks which aggregate local features into more representative features. Muqeet \emph{et al}.\cite{Muqeet_ECCV_2020} make ESA blocks more efficient by employing dilated convolutions. 
	Zhao \emph{et al}. \cite{Zhao_ECCVW_2020} explores an effective pixel attention scheme that learns attention coefficients for all pixels.
	
	\begin{figure*}[t]
		\centerline
		{\includegraphics[width=2.05\columnwidth]{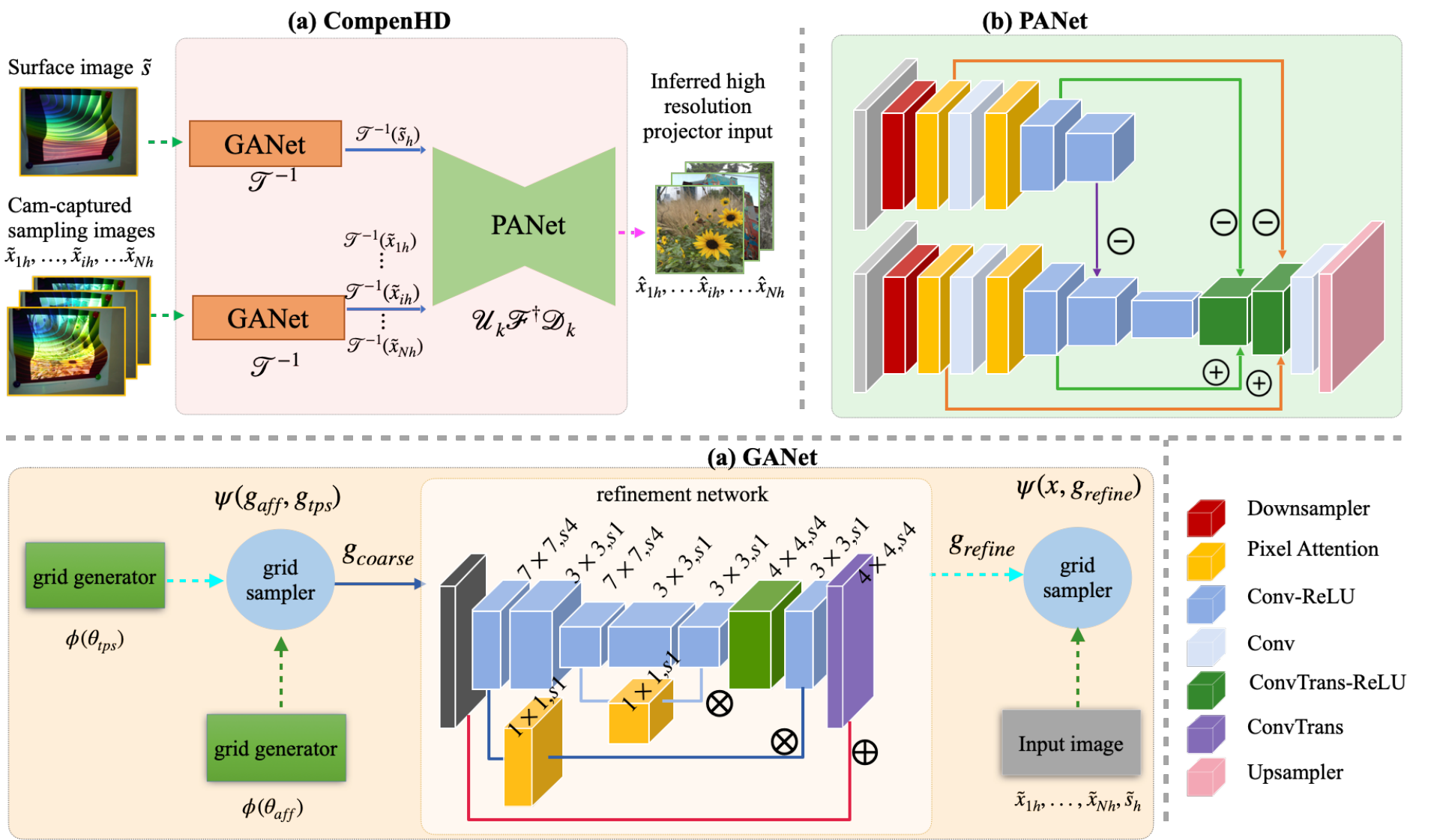}}		
		\caption{An overview of our \textbf{CompenHR}. (a) During training, \textbf{CompenHR} takes the captured surface image $\tilde{\bm{s}}$ and sampling images $\tilde{\bm{x}}_{h,1}, \dots, \tilde{\bm{x}}_{h,i}, \dots, \tilde{\bm{x}}_{h,N}$ as input, and outputs the inferred high-resolution projector input.  It consists of two subnets: \textbf{GANet} and \textbf{PANet}. (b) \textbf{GANet} aims to warp the input camera-captured image to the projector's canonical frontal view. It uses a coarse-to-fine architecture that integrates two grid generators, two grid samplers, and a novel refinement network. (c) \textbf{PANet} is applied to compensate for the warped image. It incorporates a siamese encoder with a downsampler and a decoder with an upsampler. }
		\label{framework}
		\vspace{-1.0em}
	\end{figure*}
	\section{Deep High-resolution Projector Compensation}
	\subsection{Problem formulation}
	\subsubsection{Projector compensation}
	Our full projector compensation system consists of a pair of uncalibrated high-resolution projector and camera as well as a fixed non-planar textured surface. Let the input image be $\bm{x}_h$ ($h$ stands for high-resolution), and the function that geometrically warps a high-resolution input image to the camera view be $\mathcal{T}$, and the photometric function that transforms the high-resolution warped image to the camera-captured image be $\mathcal{F}$, then the image $\tilde{\bm{x}}_h$ captured by camera\footnote{Following \cite{Huang_PAMI_2021}, we use $\tilde{~}$ for the camera-captured image.} can be formulated as:
	\begin{equation}
		\label{x_h}
		\tilde{\bm{x}}_h =\mathcal{T} (\mathcal{F}(\bm{x}_h;\bm{l},\bm{s}))
	\end{equation}
	where $\bm{l}$ stands for the environment lighting and $\bm{s}$ stands for the surface reflection parameters.
	
	The purpose of full projector compensation is to find the high-resolution compensation image $\bm{x}^*_h$, so that the camera-captured projection $\tilde{\bm{x}}^*_h$ is close to the ideal viewer-perceived image $\bm{x}'_h$:
	\begin{equation}
		\label{x*}
		\tilde{\bm{x}}^*_h =\mathcal{T} (\mathcal{F}(\bm{x}^*_h;\bm{l},\bm{s})) \approx \bm{x}'_h
	\end{equation}
	
	We assume that $\bm{s}$ and $\bm{l}$ are implicitly captured by the camera-captured surface image $\tilde{\bm{s}}$, then the compensation process can be formulated as:
	\begin{equation}
		\label{x*2}
		\bm{x}^*_h =\mathcal{F}^{\dag}(\mathcal{T}^{-1}(\bm{x}'_h);\mathcal{T}^{-1}(\tilde{\bm{s}}))
	\end{equation}
	
	\subsubsection{Compensation with a sampling scheme}
	For high-resolution setups, directly learning \equref{x*2} using deep neural networks is impractical, due to the high memory consumption and training time.
	To address this issue, we propose a more memory and time-efficient method with a novel sampling scheme below.
	
	Let the low-resolution version of $\bm{x}_h$ be $\bm{x}_l$ and plug it into \equref{x*2}, projector compensation for low-resolution input can be formulated as:
	\begin{equation}
		\label{x*l}
		\bm{x}^*_l =\mathcal{F}^{\dag}(\mathcal{T}^{-1}(\bm{x}'_l);\mathcal{T}^{-1}({\tilde{\bm{s}}_l}))
	\end{equation}
	
	Clearly, $\bm{x}'_l$ can be easily obtained by sampling $\bm{x}'_h$.  Define $\downarrow$ and $\uparrow$ as downsampling and upsampling operations, respectively, and let $k\in\{1,2,3,\dots,M\}$ be the scale factor, 
	then, according to \equref{x*2} $\bm{x}^*_h$ is given by:
	\begin{equation}
		\label{x*h2}
		\bm{x}^*_h =(\mathcal{F}^{\dag}(\mathcal{T}^{-1}(\bm{x}'_h\downarrow_k);\mathcal{T}^{-1}({\tilde{\bm{s}}_h\downarrow_k})))\uparrow_k
	\end{equation}
	where $\downarrow_k$ reduces the dimension of the image by $1/k$. This allows us to perform full compensation on the low-resolution images $\bm{x}'_l$ and ${\tilde{\bm{s}}_l}$ rather than $\bm{x}'_h$ and ${\tilde{\bm{s}}_h}$. Finally, the compensated high-resolution image $\bm{x}^*_h$ is reconstructed from $\bm{x}^*_l$ by an upsampling operation $\uparrow_k$. However, reconstructing high-resolution images from low-resolution ones is an ill-posed problem \cite{Kong_CVPRW_2022}. Thus, to preserve more information from $\bm{x}'_h$ and ${\tilde{\bm{s}}_h}$, we employ pixel unshuffle $\mathcal{D}_k$ and pixel shuffle $\mathcal{U}_k$ operations instead of $\downarrow_k$ and $\uparrow_k$, and \equref{x*2} becomes
	\begin{equation}
		\label{x*h3}
		\bm{x}^*_h =\mathcal{U}_k\Big( \mathcal{F}^{\dag}\big(\mathcal{T}^{-1}(\mathcal{D}_k(\bm{x}'_h)); \mathcal{T}^{-1}(\mathcal{D}_k(\tilde{\bm{s}}_h)) \big) \Big)
	\end{equation}
	
	Note that $\mathcal{T}$ performs image warping and the most intensive computation is performed in photometric compensation, thus we only need to perform pixel shuffle on the geometrically corrected image, therefore we swap the positions of $\mathcal{D}_k$ and $\mathcal{T}^{-1}$.
	\begin{equation}
		\label{x*h4}
		\bm{x}^*_h =\mathcal{U}_k\Big( \mathcal{F}^{\dag}\big( \mathcal{D}_k(\mathcal{T}^{-1}(\bm{x}'_h)); \mathcal{D}_k(\mathcal{T}^{-1}(\tilde{\bm{s}}_h))\big)\Big)
	\end{equation}
	
	We model the above equation using a deep neural network named CompenHR {\small$ \pi^{\dagger}_{\vec{\theta}}(\cdot, \cdot) \equiv \mathcal{U}_k\Big(\F^{\dagger}\big(\mathcal{D}_k(\T^{-1}(\cdot)); \mathcal{D}_k(\T^{-1}(\cdot))\big)\Big)$} for conciseness, where {\small$ \vec{\theta}= \{\vec{\theta}_\F, \vec{\theta}_\T\} $} are CompenHR's learnable parameters. Clearly, it can be trained using image pairs like {\small$\{\bm{x}^*_{h,i}, \bm{x}'_{h,i}\}$} and a captured surface image $\tilde{\bm{s}}_h$. However, the ground truth of $ \bm{x}^*_h $ is hard to obtain. Therefore, following \cite{Huang_PAMI_2021} we generate a surrogate training set {\small$\mathcal{X}=\{(\tilde{\bm{x}}_{h,i}, \bm{x}_{h,i})\}_{i=1}^N $} by projecting the sampling images $ \bm{x}_{h,i} $ and capturing their projections $\tilde{\bm{x}}_{h,i}$, then CompenHR can be trained by
	\begin{equation}
		\vec{\theta} = \argmin_{\vec{\theta}'}\sum_i\mathcal{L}\big(\hat{\bm{x}}_{h,i}=\pi^{\dagger}_{\vec{\theta}'}(\tilde{\bm{x}}_{h,i}; \tilde{\bm{s}}_h), \ \bm{x}_{h,i}\big)
	\end{equation}
	
	
	In our approach, we define the loss function $\mathcal{L}$ using a combination of pixel-wise ${l}_1$, ${l}_2$ and structural similarity (SSIM)\cite{Wang_TIP_2004}:
	\begin{equation}
		\label{l}
		\mathcal{L} = \mathcal{L}_{l_1} +  \mathcal{L}_{l_2} +  \mathcal{L}_{\rm ssim}
	\end{equation}
	
	
	\subsection{Network design}
	Based on \equref{x*h4}, our CompenHR integrates two subnets \textbf{GANet} and \textbf{PANet}, which model $\mathcal{T}^{-1}$ and the combination of $\mathcal{U}_k$, $\mathcal{F}^{\dag}$ and $\mathcal{D}_k$, respectively. 
	The architecture is shown in \figref{framework}(a). It takes a surface image $\tilde{\bm{s}}_h$ and some captured sampling images $\tilde{\bm{x}}_{h,i}$ of resolution $1920\times 1080$ as input, and then generates the inferred projector input/compensation images with a resolution of $1024\times 1024$. Next, we will introduce the subnets in detail.
	\subsubsection{GANet }
Our GANet is inspired by WarpingNet \cite{Huang_PAMI_2021}, a coarse-to-fine architecture for geometric correction. 	
As shown in \figref{framework}(b), GANet consists of two grid generators, two grid samplers, and a grid refinement network. Let $\vec{\theta}_{\rm aff}\in \mathcal{R}^{2\times3}$ be the learnable parameters of the affine matrix used to roughly warp $\tilde{\bm{x}}_h$ and $\tilde{\bm{s}}_h$ to the front view, $\vec{\theta}_{\rm tps}$ be the learnable parameters of thin plate spline (TPS) \cite{Donato_ECCV_2002} with five control points used to roughly model the nonlinear warping from the affine warped image to the desired view. GANet employs grid generators $\phi({\vec{\theta}_{\rm aff}})$ and $\phi({\vec{\theta}_{\rm tps}})$ to generate affine grid $\bm{g}_{\rm aff}$
and TPS grid $\bm{g}_{\rm tps}$, and then injects them into the first grid sampler $\psi(\bm{g}_{\rm aff},\bm{g}_{\rm tps})$ that samples 2D coordinates using bilinear interpolations. 
This process can be summarized as:
\begin{equation}
	\label{ganet}
	\bm{g}_{\rm coarse} =\psi(\phi(\vec{\theta}_{\rm aff}),\phi(\vec{\theta}_{\rm tps}))
\end{equation}

Then, we design a neural network $\mathcal{W}$ to further refine the coarse grid $\bm{g}_{\rm coarse}$:
\begin{equation}
	\label{ganet2}
	\bm{g}_{\rm refine}= \mathcal{W}(\bm{g}_{\rm coarse})
\end{equation}
The refinement network contains six convolutional layers followed by ReLU activation and two transpose convolutional layers. To extract useful information from a large input coarse grid efficiently, we use the first and third convolutional layers to downsample large-scale feature maps, and others to extract multi-level features. Then we place two transposed convolutional layers to upsample feature maps and generate the refined output. 
The detailed parameters are listed in \figref{framework}(b).
In addition, we employ the attention module that consists of a $1\times 1$ convolutional layer followed by sigmoid activation for efficient feature extraction. This strategy brings better performance on the geometric correction compared with \cite{Huang_PAMI_2021}.

After refining the grid, the second grid sampler is used for warping the input image using the finer sampling grid $\bm{g}_{\rm refine}$.
\begin{equation}
	\label{ganet3}
	\mathcal{T}^{-1}(\tilde{\bm{x}}_h) = \psi(\tilde{\bm{x}}_h,\bm{g}_{\rm refine})
\end{equation}

\subsubsection{PANet } 

{In PANet, we model $\mathcal{F}^{\dag}$ with a combination of an encoder and a decoder, and model the downsample/upsample-like operations $\mathcal{U}_k$/$\mathcal{D}_k$ with shuffle/unshuffle operations. }

As shown in \figref{framework}(c), to reduce computation cost and memory usage, PANet employs a pixel unshuffle operation $\mathcal{D}_k$ instead of spatial bilinear interpolation. It reshapes the input $\tilde{\bm{x}}_h \in \mathcal{R}^{1024\times 1024\times 3}$ and $\tilde{\bm{s}}_h \in \mathcal{R}^{1024\times 1024\times 3}$ to the first feature maps $\tilde{\bm{M}}_x^0\in \mathcal{R}^{256\times 256 \times 48}$ and $\tilde{\bm{M}}_s^0\in \mathcal{R}^{256\times 256 \times 48}$, without losing pixel information. Benefiting from it, useful information from the original high-resolution image can be preserved for extracting subsequent features. 

The photometric compensation function $\mathcal{F}^{\dag}$ is modeled by an encoder and a decoder.
The encoder is a siamese network with shared weights. Each branch stacks three convolutional layers, two of which are followed by ReLU activation. 
The decoder extracts and upsamples multi-level feature maps from the difference between surface feature maps and captured image feature maps. It consists of three convolutional layers and two transposed convolutional layers. The detailed parameters are noted 
in \figref{framework}(c). Besides, two skip convolution connections are used to capture interaction among multi-level information. The first skip convolution connection (yellow line in \figref{framework}(c)) consists of a $1\times1$ convolutional layer and two $3\times3$ convolutional layers. The second skip convolution connection (green line in \figref{framework}(c)) consists of a $1\times1$ convolutional layer and a $3\times3$ convolutional layer.
The stride of all convolutional layers is 1.

After generating the compensated feature maps, we employ a pixel shuffle operation $\mathcal{U}_{k}$ to recover the high-resolution image. It reshapes the multi-channel output of decoder $\tilde{\bm{M}}_n\in \mathcal{R}^{256\times 256 \times 48}$ to $\tilde{\bm{x}}_h\in \mathcal{R}^{1024\times 1024\times 3}$.
The use of pixel unshuffle and shuffle operations reduces memory usage and time computation, but it may also lead to lower precision. To alleviate this issue, 
we place two pixel attention modules \cite{Zhao_ECCVW_2020} (the yellow box in \figref{framework}(c)) after the unshuffle layer and the first convolutional layer to preserve more important information from the reshaped image and the low-level feature maps. This pixel attention module also consists of a $1\times1$ convolutional layer followed by sigmoid activation. Furthermore, it operates on the input directly by using a skip connection. Denote the input feature map as $\bm{M}\in \mathcal{R}^{C\times H \times W}$, a $1\times 1$ convolution operation as $\mathcal{C}$, the Sigmoid function as $\sigma$, respectively, then the output of the pixel attention layer is given by:
\begin{equation}
\label{cmp_att}
\bm{M}'= \sigma(\mathcal{C}(\bm{M}))\otimes \bm{M}
\end{equation}
where $\otimes$ is the element-wise multiplication.
Owing to the unshuffle/shuffle operations and the attention mechanism, the memory and time efficiency of PANet is significantly improved with little accuracy degradation.

\subsection{Training details}
We implement CompenHR using PyTorch \cite{Paszke_2019} and optimize parameters using Adam optimizer \cite{Kingma_2014}.  The model is trained on an Nvidia GeForce 1080 GPU with $2000$ iterations.

For CompenHR, the initial learning rate is set to $10^{-3}$ and is decayed by a factor of $5$ for every $1500$ iteration. For parameter initialization, the weights of the refinement network in GANet are initialized using a normal distribution with a mean of 0 and a standard deviation of 1; the weights of PANet are initialized using He’s method \cite{He_ICCV_2015}. The batch size is set to 4 for all experiments.
\section{Benchmark}
To evaluate the compensation methods for high-resolution projectors, we build a benchmark dataset with high-resolution image pairs following~\cite{Huang_PAMI_2021}.

\subsection{System configuration}
Our projector compensation system consists of a Sony $\alpha 6400$ camera and an EPSON {CB-X05} projector whose resolutions are set to $1920\times1080$ and {$1024\times768$}  respectively. 
An Elgato Cam Link 4K video capture card is used to capture the camera frames. 

The projector is placed about 1 meter in front of the surface, and the camera is placed within a range of 0.3 to 1 meter around the projector. In each setup, the camera settings such as exposure, focus, and white balance are adjusted manually based on the ambient light and surface material, and fixed during each setup data capturing. 

\subsection{{Datasets}}

\subsubsection{{Real data}}
To the best of our knowledge, there is no public high-resolution dataset for quantitative evaluation. Thus we construct a real dataset with {24} setups, {and 5 of them have  specular surfaces.} For each setup, at least one of the ambient lighting, camera parameters, non-planar projector surface, etc. is different. 

\begin{figure}[!t]
\centerline
{\includegraphics[width=\columnwidth]{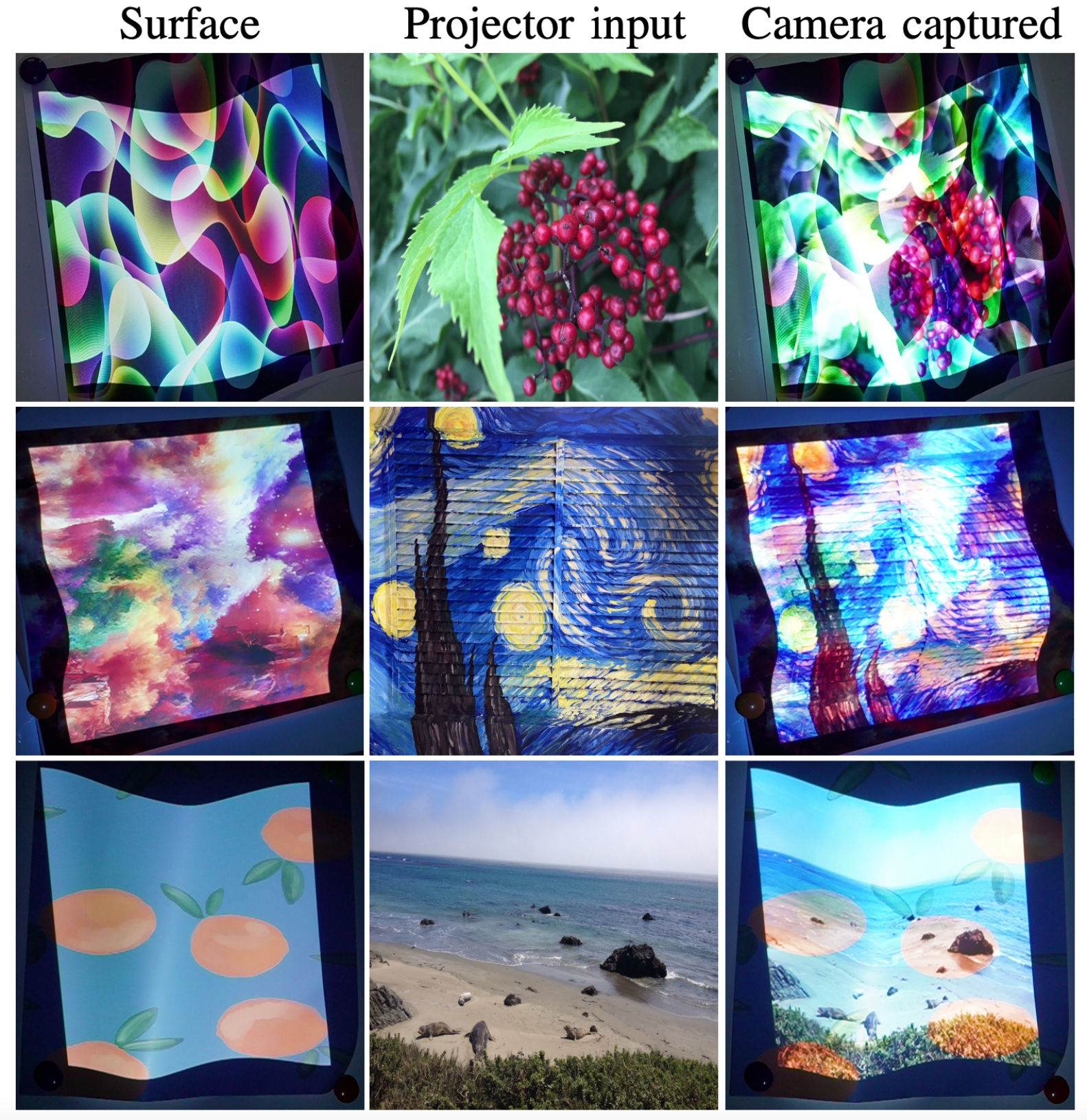}}
\caption{Samples of the real dataset. From left to right: textured surfaces, projector input images, and camera-captured projections. 
}
\label{dataset}
\vspace{-1.0em}
\end{figure}

We collect $N=700$ colorful high-resolution ($1920\times1080$ or higher) images taken in real life and resize them to $1024\times 1024$ as projector input. 
During data collection, all these images and a gray image are projected to the projection surface and captured by the camera. Thus, the sets consist of image pairs $\mathcal{X}_{\rm train}=\{(\tilde{\bm{x}}_{h,i},\bm{x}_{h,i} )| i = 1\dots N_{\rm train}\}$, $\mathcal{Y}=\{(\tilde{\bm{y}}_{h,i},\bm{y}_{h,i} )| i = 1\dots N_{\rm test}\}$ and the uncompensated surface $\tilde{\bm{s}}_{h}$. Among them, $N_{\rm train} = 500$ image pairs are for training and $N_{\rm test} = 200$ for testing. 
\figref{dataset} shows three image pairs with different setups and surfaces.

\subsubsection{{Synthetic data}}
{
Following \cite{Huang_PAMI_2021}, to improve the practicability of our full compensation method, we build a synthetic high-resolution dataset to pre-training the photometric compensation module by rendering 100 setups with different projector-camera-surface poses, materials, exposures, and lightings in Blender\cite{Blender_2020}.
We use 100 surface patterns provided by \cite{Huang_PAMI_2021} and 500 projected sampling images in this synthetic dataset are selected from DIV2K training dataset\cite{Agustsson_CVPRW_2017} and resized to $1024\times1024$.}

\subsection{Metrics}
We use the surrogate evaluation protocol presented in \cite{Huang_PAMI_2021} for quantitative comparisons, and four metrics are used: PSNR, RMSE, SSIM, and $\Delta E$ (CIE standard for perceptual color differences\cite{Sharma2005TheCC}).

\section{Experiments}\label{sec:exp}
In experiments, the proposed CompenHR is trained on our high-resolution full compensation dataset with 500 images and tested with 200 images in each setup, and the final results are averaged over K = 19 setups.

\subsection{Comparison with state-of-the-arts}
We compare our CompenHR with an end-to-end trainable method CompenNeSt++. The original CompenNeSt++ is proposed as the solution for low-resolution input, we trained it with both high-resolution and low-resolution image pairs, and we name the two baselines \emph{CmpSt(HR)} and \emph{CmpSt(LR)}, respectively. Besides, an intuitive way to reconstruct high-resolution images is using super-resolution. Thus, we also use CompenNeSt++ to generate low-resolution compensation images and then reconstruct the corresponding high-resolution images by bicubic interpolation, and we name this method \emph{CmpSt(Bi)}. We also combine CompenNeSt++ with the state-of-the-art deep learning-based super-resolution method named Residual Local Feature Network (RLFN) \cite{Kong_CVPRW_2022}. We pre-train it with DIV2K training dataset \cite{Agustsson_CVPRW_2017} and use it as an upsampler without fine-tuning, we name this baseline as \emph{CmpSt(SR)}.

We train \emph{CompenHR} and \emph{CmpSt(HR)} using high-resolution image pairs ($1024\times1024$) and then train \emph{CmpSt(LR)} using the low-resolution image pairs ($256\times256$).
In testing, all methods' input and output resolutions are set to $1024\times 1024$, but the intermediate resolutions of the two two-step methods (\emph{CmpSt(Bi)} and \emph{CmpSt(SR)}) are $256\times 256$.	
The quantitative comparisons are shown in \tabref{t-sota-ave}. To verify the efficiency of algorithms, in \tabref{t-sota-time} we further compare the time and memory consumption of \emph{CmpSt(LR)}, \emph{CmpSt(HR)}, and our \emph{CompenHR} during training. We ignore the consumption during testing since it is negligible compared with the training phase. \emph{CmpSt(Bi)} and \emph{CmpSt(SR)} share the same trained module \emph{CmpSt(LR)}, thus they have the same parameters, FLOPS, training time, and memory consumption. 

\begin{table}[!ht]
\caption{Quantitative comparison of full compensation algorithm on image quality. Results are averaged over 19 different setups. 
}

\centering
\begin{tabular}{lcccc}
	\toprule[0.5mm]
	
	Model & PSNR & RMSE & SSIM &$ \Delta E$\\ \midrule[0.5mm]
	CmpSt(HR) &20.5521	& 0.1627	& 0.5980	& 7.6158\\ 
	CmpSt(LR) &17.7584	& 0.2259	& 0.4654	& 9.9362 \\ 
	CmpSt(Bi) & 19.9153	& 0.1758	& 0.5499	& 8.1453\\ 
	CmpSt(SR) & 19.9051	& 0.1758	& 0.5497	& 8.1797\\
	CompenHR & 		\textbf{20.9496}	&\textbf{0.1554}	& \textbf{0.6012} & \textbf{7.5901}\\   \midrule[0.3mm]
	Uncompensated &11.4813	& 0.4676	& 0.2384	& 21.7226 \\ 
	\bottomrule[0.5mm]
\end{tabular} 
\label{t-sota-ave}
\end{table}

\begin{table}[!ht]
\caption{{Quantitative comparison of full compensation algorithm on image quality. Results are averaged over 5 different setups with specular highlight surfaces. }
}

\centering
\begin{tabular}{lcccc}
	\toprule[0.5mm]
	{Model} & PSNR & RMSE & SSIM &$ \Delta E$\\ \midrule[0.5mm]
	{CmpSt(HR)} &20.6376	&0.1614	&0.6049	&7.3914\\ 
	{CmpSt(LR)} &18.3643	&0.2104	&0.4772	&9.1279 \\ 
	{CmpSt(Bi)} &20.1215	& 0.1710	&0.5514	&7.8007\\ 
	{CmpSt(SR)} & 20.1094	&0.1712	&0.5513	&7.8371\\
	{CompenHR} & \textbf{21.0135	}	&\textbf{0.1544}	& \textbf{0.6094} & \textbf{7.3994}\\   \midrule[0.3mm]
	{Uncompensated} &11.4566	&0.4695	&0.2327&	20.5489 \\ 
	\bottomrule[0.5mm]
\end{tabular} 
\label{t-sota-ave2}
\end{table}

\begin{table}[!ht]
\caption{Quantitative comparison of full compensation algorithm on the amount of computation.{ Results are averaged over 19 different setups. }
}

\centering
\resizebox{0.48\textwidth}{!}
{
	\begin{tabular}{lcccc}
		\toprule[0.5mm]
		Model & Params & FLOPS(G) & Mem.(M)  & Time(s)\\ \midrule[0.5mm]
		CmpSt(HR) & \textbf{833,145} & 1645.5460 & 10003  & {6604}\\ 
		CmpSt(LR)& \textbf{833,145}&\textbf{102.8466} & \textbf{1779}  &\textbf{367.79}  \\ 
		CompenHR & 1327,586 & 364.1476& 4927 & 1487.79 \\ 		\bottomrule[0.5mm]
	\end{tabular} 
}
\label{t-sota-time}
\end{table}

\begin{figure*}[t]
\centering
\includegraphics[width=2.1\columnwidth]{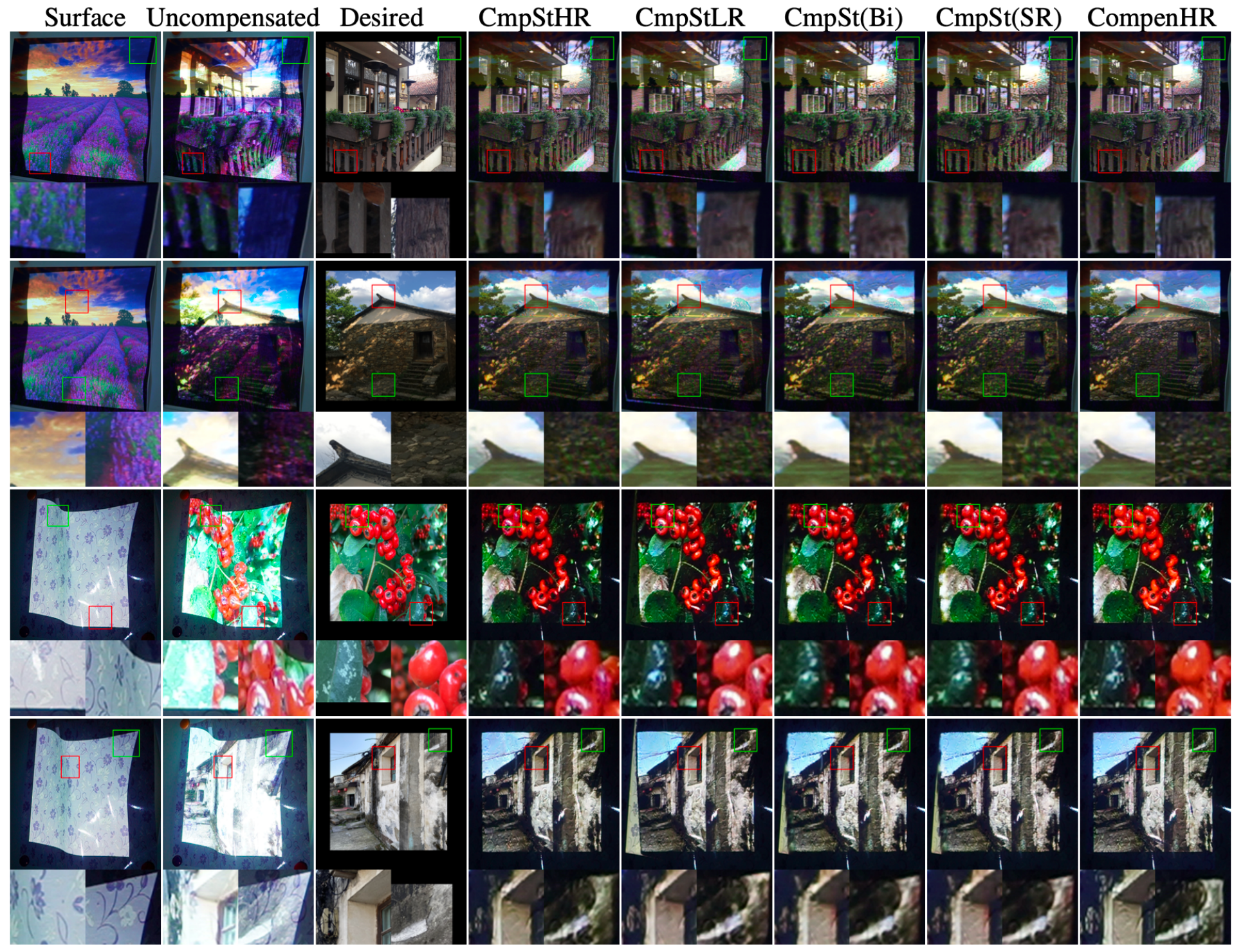}\\
\caption{Qualitative comparison on real camera-captured compensation. From top to bottom: surface, uncompensated, desired image, \emph{CmpSt(HR)}, \emph{CmpSt(LR)}, \emph{CmpSt(Bi)}, \emph{CmpSt(SR)}, and our \emph{CompenHR}. See high-resolution figures in the supplementary.}
\label{f-res-sota}
\vspace{-0.5em}
\end{figure*}

In \tabref{t-sota-ave}, {\tabref{t-sota-ave2}} and \tabref{t-sota-time}, comparing our \emph{CompenHR} with \emph{CmpSt(Bi)} and \emph{CmpSt(SR)}, because our \emph{CompenHR} is trained with high-resolution images, it has a higher FLOPS, number of parameters and memory usage in training. But clearly, it also has a better compensation quality than others in testing
Besides, compared with \emph{CmpSt(HR)}, which also uses high-resolution images for training, \emph{CompenHR} achieves a little better quality and trains much faster. Benefiting from the usage of shuffle/unshuffle operations, the feature map size is reduced to a quarter of the original sizes so that our \emph{CompenHR} consumes less computation and memory than \emph{CmpSt(HR)}.
\figref{f-res-sota} shows qualitative comparisons of all methods. For all samples, the color and brightness of \emph{CompenHR} are better than the others, and the results of \emph{CompenHR} and \emph{CmpHR} are sharper than the other three methods. In particular, the left two columns also show that \emph{CompenHR}'s geometric correction is more accurate than the others.
These results further demonstrate that preserving the input image resolution has a significant influence on the quality of compensation images. Methods trained with high-resolution images learn more details than low-resolution ones.
{In particular, these methods can handle slight specular highlights to some extent but do not work well on the area with strong specular reflection.}

\subsection{Ablation study}
{In this section, we first validate the proposed CompenHR using different numbers of training datasets, and then} considering the major ingredients in CompenHR: (1) a novel GANet with sampling grid refinement; (2) shuffle/unshuffle based sampling operations for both input and output of PANet; (3) pixel attention mechanisms for important feature extraction; (4) the loss function combining ${l}_1$, ${l}_2$ and $SSIM$, we explore the effectiveness of our sampling grid refinement network, attention mechanism, and loss function.

\subsubsection{{The effect of the number of training images }}
{To validate the practicability of the proposed method further, we train our network with different numbers of training images and evaluate them on our real dataset with 19 setups. The results are reported in \tabref{t-res-shuffle-trainingnumber}. The image quality of the default CompenHR using 48 training images is much higher than that using 8 training images, and then the image quality improves slightly as the number of training datasets increases.}

{Furthermore, following \cite{Huang_PAMI_2021}, we also pre-train CompenHR using the proposed synthetic dataset with 100 setups, and then \textit{fine-tune it using only 8 images with $1000$ iterations}. The initial learning rate for fine-tuning is set to $10^{-3}$ and is decayed by a factor of $5$ for every $600$ iteration. 
In \tabref{t-res-shuffle-pretrain}, the pre-trained CompenHR outperforms the default CompenHR. }
\begin{table}[h]
\centering
\caption{
	{ Quantitative comparisons of CompenHR with different numbers of training datasets. Results are averaged over 19 different setups. } 
}
\label{t-res-shuffle-trainingnumber}
\begin{tabular}{lcccc}
	\toprule[0.5mm]
	Model(-\#Train) & PSNR & RMSE & SSIM & $\Delta E$ \\ \midrule[0.5mm]
	{CompenHR-8 }&19.3464	&0.1886	&0.5137	&9.2381 \\ 			
	{CompenHR-48} &  20.7636 &	0.1587	& 0.5938& 	7.8099 \\
	{CompenHR-125} &  20.7505 &	0.1590	&0.5944	& 8.0303  \\	
	{CompenHR-250} & 20.7871	& 0.1584	& 0.5988 &	7.7396 \\	
	CompenHR-500 & \textbf{20.9496}	&\textbf{0.1554}	& \textbf{0.6012} & \textbf{7.5901}\\ 
	\bottomrule[0.5mm]
\end{tabular}
\end{table}

\begin{table}[h]
\centering
\caption{
	{ Quantitative comparisons between the default CompenHR and the pre-trained CompenHR. Results are averaged over 19 different setups. } 
}
\label{t-res-shuffle-pretrain}
\resizebox{0.48\textwidth}{!}
{
	\begin{tabular}{lcccc}
		\toprule[0.5mm]
		Model(-\#Train) & PSNR & RMSE & SSIM & $\Delta E$ \\ \midrule[0.5mm]
		{CompenHR-8} &  19.3246	& 0.1901	& 0.5325	& 9.4281\\
		{CompenHR-pretrain-8} &  \textbf{19.6698}	& \textbf{0.1808}	& \textbf{0.5516}	&\textbf{8.9413 } \\	
		\bottomrule[0.5mm]
	\end{tabular}
}
\end{table}
\begin{table}[h]
\centering
\caption{Quantitative comparisons of CompenHR with different sampling grid refinement networks.}
\label{t-res-shuffle-ganet}
\begin{tabular}{lcccc}
	\toprule[0.5mm]
	Model & PSNR & RMSE & SSIM & $\Delta E$ \\ \midrule[0.5mm]
	{CmpHR(WPGA) }&20.4837	&0.1640	&0.5741	&7.8553\\ 			
	{CmpHR w/o $r_1,r_2$} &  20.6765	& 0.1606& 	0.5846& 	7.6971  \\
	CompenHR &  \textbf{20.9496}	&\textbf{0.1554}	& \textbf{0.6012} & \textbf{7.5901}\\ 
	\bottomrule[0.5mm]
\end{tabular}
\end{table}

\subsubsection{Effectiveness of the refinement network in GANet}
\begin{figure*}[!ht]
\centering
\includegraphics[width=1.8\columnwidth]{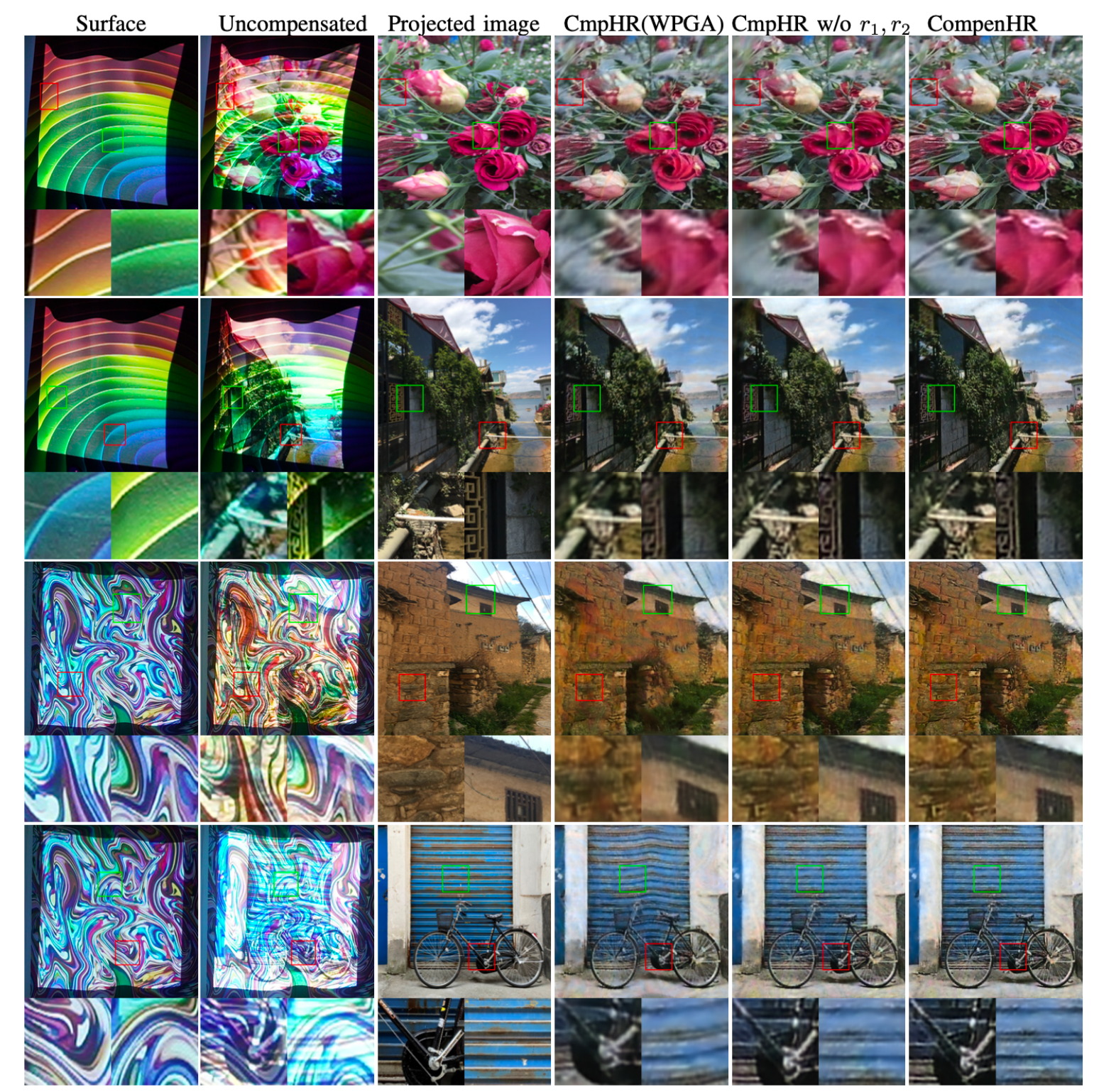}
\caption{Qualitative comparison of models with different geometric correction methods. From top to bottom: surface, uncompensated, projected image, \emph{CmpHR(WPGA)}, \emph{CmpHR w/o $r_1,r_2$} and \emph{CompenHR}. See high-resolution figures in the supplementary.}
\label{f-res-shuffle-ga}
\vspace{-1.2em}
\end{figure*}

To show the effectiveness of GANet, we replace GANet with WarpingNet in \cite{Huang_PAMI_2021} and name the compensation model \emph{CmpHR(WPGA)}. As the sampling grid refinement network contains two attention modules (yellow blocks in \figref{framework}(b)), to further explore the role of this architecture, we also compare \emph{CompenHR} with the model without $r_1$ and $r_2$ (short for \emph{CmpHR w/o $r_1,r_2$}). The quantitative comparisons in \tabref{t-res-shuffle-ganet} show that \emph{CompenHR} and \emph{CmpHR w/o $r_1,r_2$} outperform \emph{CmpHR(WPGA)} on all metrics. 
Additionally, the fact that the image quality of \emph{CompenHR} is better than \emph{CmpHR w/o $r_1,r_2$} also indicates the effectiveness of the attention modules.
In \figref{f-res-shuffle-ga}, \emph{CompenHR} generates the sharpest images for all examples. Besides, in the two right-most columns, \emph{CmpHR(WPGA)} cannot work well for the cluttered surface texture, as it can not warp the image correctly. 

\begin{table}[!h]
\centering
\caption{Quantitative comparisons of CompenHR with different pixel attention layers in PANet.}
\label{t-res-shuffle-pa}
\begin{tabular}{lcccc}
	\toprule[0.5mm]
	Model & PSNR & RMSE & SSIM & $\Delta E$ \\ \midrule[0.5mm]
	{CmpHR  w/o $p_1$} & 20.6688 &	0.1606	&0.5932	& 7.7296 \\
	{CmpHR  w/o $p_2$} & { 20.8284	}& {0.1576} & 	{0.5953}	& {7.6685} \\		
	{CmpHR w/o $p_1,p_2$} & 20.4553	&0.1646	& 0.5915 & 8.1839 \\
	CompenHR & \textbf{20.9468}	&\textbf{0.1554}	& \textbf{0.6011} & \textbf{7.5746}\\ 
	\bottomrule[0.5mm]
\end{tabular}
\vspace{-1.0em}
\end{table}

\subsubsection{Effectiveness of pixel attention blocks in PANet}
\begin{figure*}[!tb]
\centering
\includegraphics[width=2.0\columnwidth]{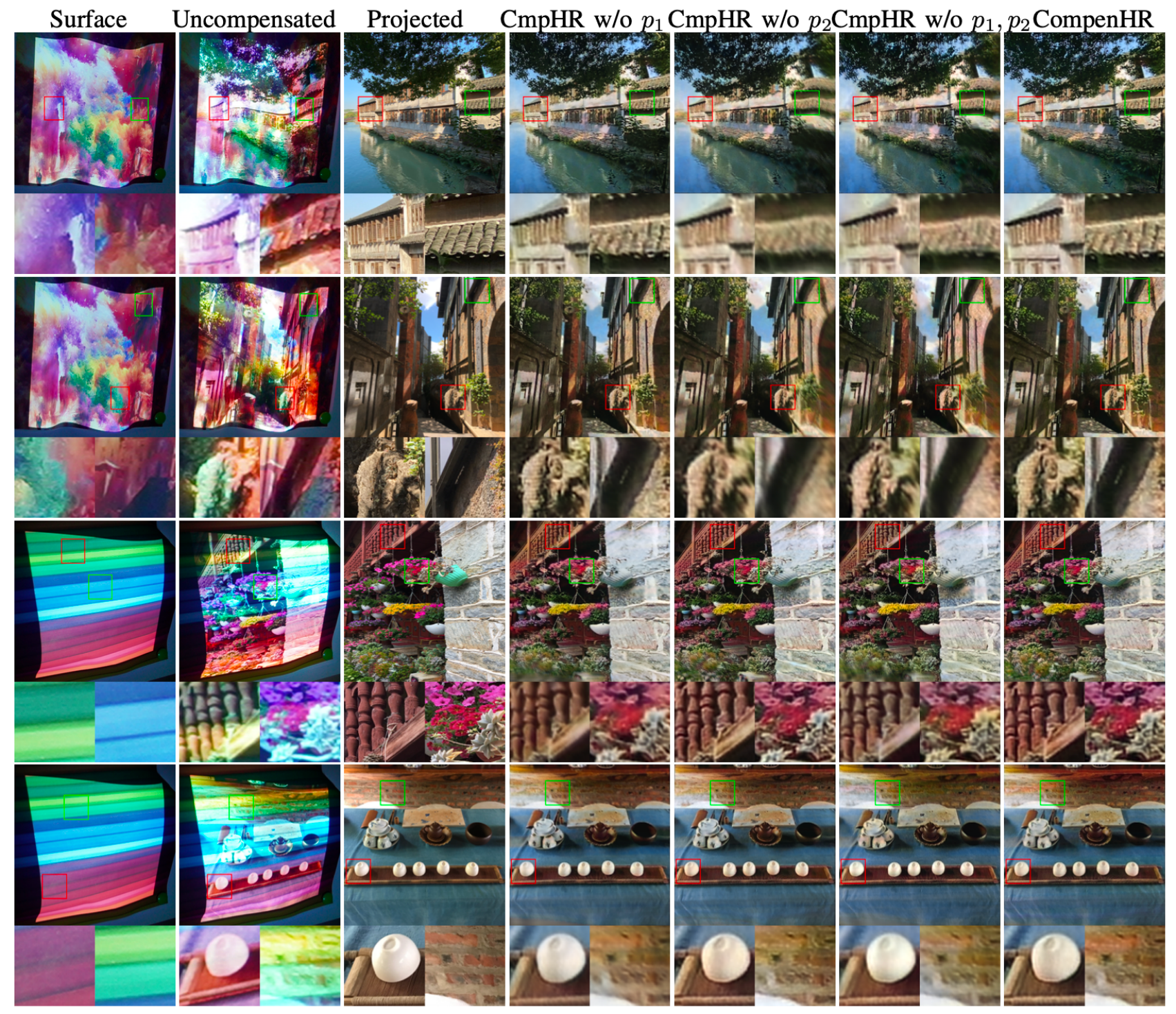}
\caption{{Qualitative comparison of models with the different number of attention layers. From top to bottom: surface, uncompensated, projected image}, \emph{CmpHR w/o $p_1$}, \emph{CmpHR w/o $p_1$}, \emph{CmpHR w/o $p_1,p_2$} and \emph{CompenHR}. {See high-resolution figures in supplementary.}}
\label{f-res-shuffle-pa}
\vspace{-1.2em}
\end{figure*}

To improve the performance of CompenHR, two attention modules are also introduced to extract key features in PANet. We explore their effectiveness in this section. After removing each pixel attention layer respectively, we build new models named \emph{CmpHR w/o $p_1$}, {\emph{CmpHR w/o $p_2$}}, and \emph{CmpHR w/o $p_1,p_2$}, respectively. 
In the comparison experiment, all methods are trained using the proposed full compensation datasets with 2,000 iterations. 
The quantitative comparisons are listed in \tabref{t-res-shuffle-pa} and the qualitative comparisons are shown in \figref{f-res-shuffle-pa}.

In the quantitative comparison, their SSIM scores are very close, while PSNR scores gradually increase, and RMSE and $\Delta E$ scores decrease with the number of pixel attention layers. In particular, the image quality of \emph{CmpHR w/o $p_2$} is slightly better than \emph{CmpHR w/o $p_1$}. In \figref{f-res-shuffle-pa}, the compensation images of \emph{CompenHR} have the closest color and detail to the desired effects. Besides, the detail of images generated by \emph{CompenHR} is close to \emph{CmpHR w/o $p_2$}, and is slightly better than \emph{CmpHR w/o $p_1,p_2$}, because the first pixel attention module extracts key features from the reshaped colorful image directly and the second one further extracts key features from its linear
transformation results. Benefiting from the attention mechanism, \emph{CompenHR} preserves more image color information in photometric compensation.
The experiment demonstrates that the pixel attention mechanism with a few additional parameters can help the model achieve better performance on image color. 
But adding more attention blocks only brings a small performance improvement. Thus, we finally use two attention blocks in our method.

\subsubsection{Comparison of different loss functions}
The pixel-wise $l_1$ and $l_2$ losses are widely used to penalize the pixel errors in many image reconstruction tasks. In \cite{Huang_PAMI_2021}, Huang \emph{et al}. verify that $SSIM$ loss can be used for image compensation tasks to help recover the structural details. Therefore, we compare the performance of methods that use different combinations of these three loss functions in \tabref{t-res-loss}. 

When using three losses separately, $l_1$ loss achieves the best scores on all metrics. 
Using $l_2$ or $SSIM$ alone achieves suboptimal results in this task, while loss functions with added $SSIM$ loss achieve higher structure similarity, and loss functions with added $l_l$ loss achieve better color quality. More qualitative comparisons are listed in the supplementary material. 
The results confirm that $l_1$ loss contributes to compensating image color in this task, while $SSIM$ loss tends to recover image structural details. In addition, $l_2$ loss also helps improve the image quality slightly. As a result, we employ the combination of all three losses and achieve the best performance.

\begin{table}[!tbh]
\centering
\caption{ {Quantitative comparisons of CompenHR with different loss functions.} }
\begin{tabular}{lcccc}
	\toprule[0.5mm]
	Loss & PSNR & RMSE & SSIM & $\Delta E$\\ \midrule[0.5mm]
	$l_1$ & 20.7247	 & 0.1595& 	0.5571	& 7.7462 \\ 
	$l_2$& 20.3679	& 0.1663	& 0.5465	& 8.1671\\ 
	$SSIM$& 18.9372 &	0.1964 & 0.5471	& 11.3003\\ 
	$l_1+l_2$ &20.7946	& 0.1582& 	0.5555	& 7.7017 \\ 
	$l_1+SSIM$&20.8897	& 0.1564	& 0.5984	& 7.6284 \\ 
	$l_2+SSIM$ & 20.3312&	0.1670	&0.5917	& 8.7418 \\  
	$l_1+l_2+SSIM$ & \textbf{20.9468}	&\textbf{0.1554}	& \textbf{0.6011} & \textbf{7.5746}\\ \bottomrule[0.5mm]
\end{tabular}
\label{t-res-loss}
\end{table}
\section{{Discussion}}

{Our method improves the efficiency of the deep learning-based method and achieves competitive performance on the high-resolution compensation task, but it still has some limitations. First, our model is designed for static projector-camera systems, and in future work, we will explore online learning methods for dynamic projector compensation.}{ Second, like \cite{Huang_PAMI_2021}, our GANet does not work for surfaces with sharp edges and occlusions, and a multi-projector setup may better address this issue.}
{Third, similar to CompenNeSt++\cite{Huang_PAMI_2021}, our method can handle slight specular highlights but does not work well on the area with strong specular reflections. }

\section{Conclusion }
In high-resolution full projector compensation, memory usage and time cost increase sharply with the image resolution. This paper proposes an efficient end-to-end solution by first reformulating the full compensation problem by integrating the sampling process, then an attention-based sampling grid refinement network is designed for better geometric correction. Moreover, unshuffle/shuffle operations and pixel attention mechanisms are applied to improve quality and efficiency. Finally, a high-resolution full compensation benchmark dataset is constructed, and experiments demonstrate the advantages of the proposed method.

\acknowledgments{
This work was supported in part by Fundamental Research Funds for the Central Universities(SWU122001), the Nature Science Foundation of China(62202135), and the Open Project Program of the State Key Laboratory of Virtual Reality Technology and System, Beihang University(VRLAB2021C09). HL was not supported by any research fund.}

\bibliographystyle{abbrv-doi}

\bibliography{CAMERA-READY_202311}
\end{document}